# Semantic segmentation on multi-resolution optical and microwave data using deep learning

Jai Singla and Bakul Vaghela

*Space Applications Centre, (Indian Space Research Organization), Ahmedabad (Gujarat), India.*

E-mail: **jaisingla@gmail.com** *-Corresponding Author, Orcid Id: **0000-0003-3893-2383**

# Semantic segmentation on multi-resolution optical and microwave data using deep learning

**Abstract**

Presently, deep learning and convolutional neural networks (CNNs) are widely used in the fields of image processing, image classification, object identification and many more. In this work, we implemented convolutional neural network based modified U-Net model and VGG-UNet model to automatically identify objects from satellite imagery captured using high resolution Indian remote sensing satellites and then to pixel wise classify satellite data into various classes. In this paper, Cartosat 2S (~1m spatial resolution) datasets were used and deep learning models were implemented to detect building shapes and ships from the test datasets with an accuracy of more than 95%. In another experiment, microwave data (varied resolution) from RISAT-1 was taken as an input and ships and trees were detected with an accuracy of >96% from these datasets. For the classification of images into multiple-classes, deep learning model was trained on multispectral Cartosat images. Model generated results were then tested using ground truth. Multi-label classification results were obtained with an accuracy (IoU) of better than 95%. Total six different problems were attempted using deep learning models and IoU accuracies in the range of 85% to 98% were achieved depending on the degree of complexity.

## 1. Introduction

In the last few years, deep convolutional networks have outperformed the state of the art methods in many visual recognition tasks. The typical use of convolutional networks is on classification tasks, where the output to an image is a single class label. However, in many visual tasks, the desired output should include localization, i.e., a class label is supposed to be assigned to each pixel. In case of an image-classification/ segmentation problem, machine has to partition the image in to different segments, each of them representing different entity. We implemented Convolutional Neural Network (CNN) based deep learning model to assign a class label for each pixel in our work.

Researchers across the world had already worked on various deep learning models for object detection and classification from satellite and aerial images. Papadomanolakia et al. (2016) worked on benchmarking of deep learning frameworks for the classification of high resolution multispectral data. They evaluated performance of various CNN based models on publically available SAT-4 and SAT-6 datasets with an accuracy of more than 99%. Prathap et. al. (2018) worked on building detection from multispectral satellite imagery based on U-Net architecture and they compared their model with SpaceNet-2 wining solution. Their model outperformed other models in the case of detection of big building structures whereas for smaller building structure, further improvements are required. Ivanovsky et. al. (2019) presented research results on two convolutional neural networks (U-Net and LinkNet) for building detection on satellite images of Planet database. As per their study,

U-Net based CNN outperformed LinkNet in terms of accuracies. Mnih and Hilton detected roads in high-resolution aerial images using neural networks. Many researchers also touched upon ship detection and tree detection using satellite imageries. Zhou et. al. (2019) proposed a ship detection method for PolSAR images based on modified faster region-based convolutional neural network (Faster R-CNN). Based on probability and figure of merit (FoM) metric, their Faster R-CNN model facilitated faster and accurate detections. Delalieux et. al. (2020) implemented convolutional neural networks (CNN) to generate a detection model, which was able to locate and classify individual palms trees from aerial high-resolution RGB images. Ke et al. (2018) studied urban land use and land cover classification using novel deep learning models based on high spatial resolution imagery. They trained their ResAPP-Unet model based on five different land cover classification on WorldView-2 (WV) and WV-3 images over Beijing and achieved the accuracy of 87.1 % on WV-2 and 84% on WV-3 images respectively. Multi-label classification of Amazon satellite images with CNN approach is carried out by Gardner et. al. (2017). They demonstrated the results using different CNN models and achieved F-score of 0.91 using ResNet-50 model. Ulmas et al. (2020) implemented a modified U-Net structure for creating land cover classification mapping based on satellite imagery. They made use of the Corine land classification dataset, BigEarthNet dataset for their research work and demonstrated land cover classification results using deep learning.

In this work, different types (microwave and optical images of varied spatial resolution) of datasets were taken from Indian remote sensing satellite and objects such as buildings, ships, and tree were automatically detected using the technique of deep learning and semantic segmentation. Initially, training data banks were prepared for each of the problem separately, modified U-Net model and VGG-Unet model were trained on the training datasets with an IoU accuracy of more than 99% in more than 100 epochs and further trained model was applied on unseen datasets. After achieving good accuracies from object detection of buildings/ships/trees, multi- label pixel based segmentation was implemented on the remote sensing datasets and accuracies of better than 95% were demonstrated. This work is unique in a way because, ship detection/ tree detection/ building detection and multi-label classification on varied resolution of datasets are demonstrated using deep learning and high resolution Indian remote sensing datasets with such a good accuracy. Our machine learning model converged quickly with relatively lesser data (200 images in some of the cases) and demonstrated high accurate results.

## 2. Data sets used

Indian Space Research Organization (ISRO) has launched variety of earth observation missions in optical as well as microwave domain. In the microwave domain, RISAT series is the leading whereas in optical domain Cartosat-2S series is launched for high resolution remote sensing. In this paper, two different types of datasets (microwave varied resolution, optical high resolution) were used to automatically detect various objects. Datasets were used for solving three different problem types. Following satellites and data products were used for training and validation purpose in this study.

### a. Cartosat-2S Multispectral data

Cartosat-2S is series of high resolution satellites launched by ISRO. Satellite is capable of a long-track steering and up to across track steering from an altitude of 500 km. It has 5 days' revisit with across track steering. Spatial resolution is ~0.65m for panchromatic images and ~2 m for multi spectral images with a swath of about 10 km. It was launched in the year 2016. (ISRO website). We have used datasets over three different places namely

Ahmedabad, Jaipur and Mumbai acquired using Cartosat-2S for building detection purpose, C2S datasets near to sea area for ship detection and C2S and Carto3 tiles for tree detection. Details of the dataset for various object detections exercises are mentioned in the Table-1.

**b. Microwave data**

RISAT-1 is a state of art series of microwave remote sensing satellite carrying a synthetic aperture radar payload operating in C band, which enables imaging during day and night times. It provides 1m to 24-meter spatial resolution datasets based on the different types of acquisition modes. (ISRO website). Microwave RISAT-1 and RISAT-2B datasets are used for ship detection and tree detection from microwave images. Descriptive details about used dataset is mentioned in the Table-1.

| Satellite - Scene | Spatial resolution | Product ID | Date of acquisition | Dimensions | Purpose |
|---|---|---|---|---|---|
| Cartosat-2S Ahmedabad | ~1.6m | 16512511 | 24-Feb-2017 | 8605x8044 | Building detection |
| Cartosat-2S Jaipur | ~1.6m | 16878611 | 20-Mar-2017 | 7764x7501 | Building detection |
| Cartosat-2S Mumbai | ~1.6m | 16272511 | 18-Feb-2017 | 6764x6650 | Building detection / Ship detection |
| RISAT-1 | 18 m | 11342mpd1_s17 | 18-MAY-2014 | 7828x7886 | Ship detection |
| RISAT-1(FRS1) | 2 m | 65771sd1_s4 | 06-JUL-2013 | 13081x8843 | Tree detection |
| RISAT-2B (FINESPOT) | 0.4 m | RB2_543a_E_s1 | 12-DEC-2020 | 31064x28117 | RISAT-2B (FINESPOT) |

**Table-1 Data Products and its specifications**

**3. Methodology**

Due to increasing rate of satellites and acquired datasets, there is requirement to carry out data analytics at the faster pace and in automatic manner. With the applications of CNNs in computer visions, image processing and related domains, deep learning is showcasing promising results for classification and object detection tasks. One of the task of classification is to find out bounding box around the object in the target images. Whereas, semantic segmentation does not predict any bounding boxes around the objects instead it figures out the pixels in the entire image which belongs to a particular object. So, it does not distinguish among various instances of a single object rather all the instances of similar objects will be represented by single color values. We are explaining the details of datasets, ground truth, model architecture, optimization functions, error loss functions and evaluation metrics used in this exercise.

**3.1 Datasets & Ground Truth Preparation**

It is most challenging and tedious task to prepare input dataset for deep learning tasks. Datasets from Indian remote sensing mainly from CartoSat series and RISAT series satellites are used for training and validation purpose in all the different exercises. From the actual satellite images of different datasets, smaller tiles of 512x512 pixels were generated and corresponding ground truth was manually prepared against each feature. In the exercise, we started from image data encoded as integers in the 0–255 range, encoding grayscale values. Before we fed this data into our network, we assigned each different class a label value of 0,1,2 up to N in the ground truth mask. So, for each class we did normalization step and assigned different values from 0 to N. For example, buildings are assigned value 0 and non-buildings as value 1 for detection of buildings, similarly, ships and non-ships are assigned 0 and 1 respectively in optical and microwave datasets. For multi-label classifications, different classes like urban, water, land and tree are assigned different values of 0,1,2,3 and pixels not belonging to any of this class are labelled as 4. It is very difficult to prepare one to one ground truth for pixel level details. Training and test data were partitioned as per 80:20 ratios. For some of the exercises, we have only 160 images with dimension of 512x512 to train the model and 40 images to validate and evaluate the model. For each problem type of varied spatial resolution and sensor type, separate set of training data and ground truth is prepared whereas uniform model architecture is used to solve the problem type.

## 3.2 Model Architecture:

Usually, convolutional neural network model contains several convolutional layers, nonlinear activations, batch normalization, and pooling layers. For image classification, mapping of the spatial tensor from the convolution layers to a fixed length vector is required. To do that, fully connected layers are used, which destroy all the spatial information. Whereas for semantic segmentation spatial information has to be retained, hence no fully connected layer is used.

U-Net architecture adopts an encode-decode framework with skip connections. U-Net model is advantage because it works with few training samples and provides better performance (Ronneberger et al, 2015). U-Net architecture consists of three sections: contractions, bottleneck and expansion as shown in Figure-1 where contraction maps to down-sampling, expansion maps to up-sampling and bottleneck helps modes to learn features. Segmentation networks like U-Net usually have three main components: convolutions, down-sampling, and up-sampling layers. Down sampling is used to reduce the dimension of the input image; for example in first few convolution operations, dimensions are reduced from 512x512x1 to 256x256x64 whereas feature vectors are increased from one to 64. Convolution operations are basic filter operations applied on input layer to extract meaningful output and feature vector stands for number of filters applied on the input layer. After the down-sampling part, we have a feature vector with shape [w,h,d] where w, h and d are the width, height and depth of the feature tensor. This compressed feature vector is fed to a series of up-sampling layers. Transposed convolution is used as up sampling technique that expands the size of the images. Image is upsized from 28x28x1024 to 56x56x512 to the last layer with a filter size of 1x1. Long Skip connections are available between the layers to add extra global information for decoding.

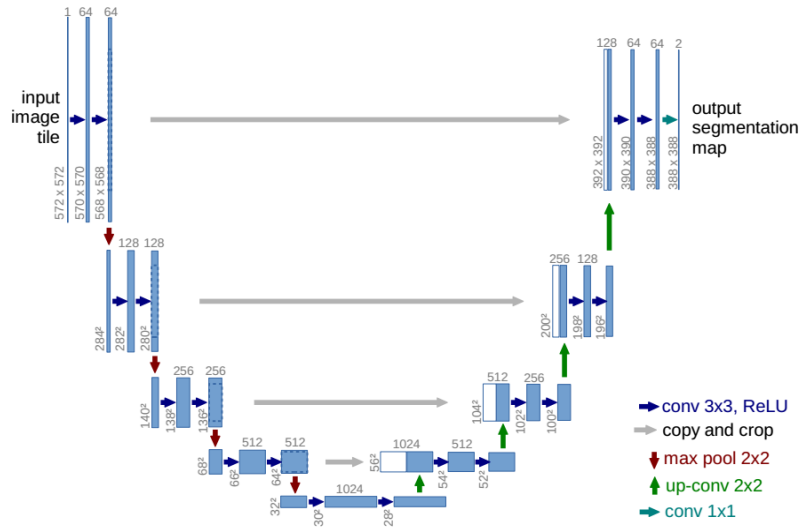

**Figure-1: Architecture of U-Net (source: Ronneberger et al.)**

**Modified U-Net architecture**

For each problem type (building detection, ship detection, tree detection and so on), separate batch of input images and their corresponding labels are used to train our deep learning model. We designed our modified U-Net for binary segmentation and VGG-UNet training models (Unet model with VGG as backbone) for multi-label segmentation problem using Keras, Python and applied it on the separate banks of two different sensors that is CartoSat (spatial resolution of ~1m), and RISAT various modes. Basic flowchart of model input, output and processing is shown in Figure-2. Initially, the actual satellite images were converted into smaller tiles of 512x512 size and corresponding ground truth was generated manually for each of the problem type separately. Ground truth was then normalized in the range of 0-N depending on the numbers of classes. Encoder- decoder based U-Net architecture with total 10 convolutional layers was used to solve all binary segmentation (building/ship/tree detection) problems mentioned in the paper. In our modified U-Net architecture, only five number of layers were used as contraction layers and five number of layers were used as expansion layers. The skip connections between the blocks were used to restore the dimensions of the image and was implemented using concatenate operation. Softmax was used as a final activation layer.

In between batch normalization, max pooling and RELU activation layers were also used. Max pooling was used to represent only the important information (calculates max. value for patches of a feature map) while down sampling. RELU activation function was used as activation function because it avoids the problem of vanishing gradient as it has maximum threshold value to infinity. Batch normalization and drop outs were used to avoid the problem of model overfitting. More than 100 epochs / iterations were used for optimum results. Model was optimized to ten layers with adaptive learning rate, batch normalization and with the use of RELU activation layer as it converged in very less time comparatively. We also made use of high computer graphics processing unit to train the model in faster manner.

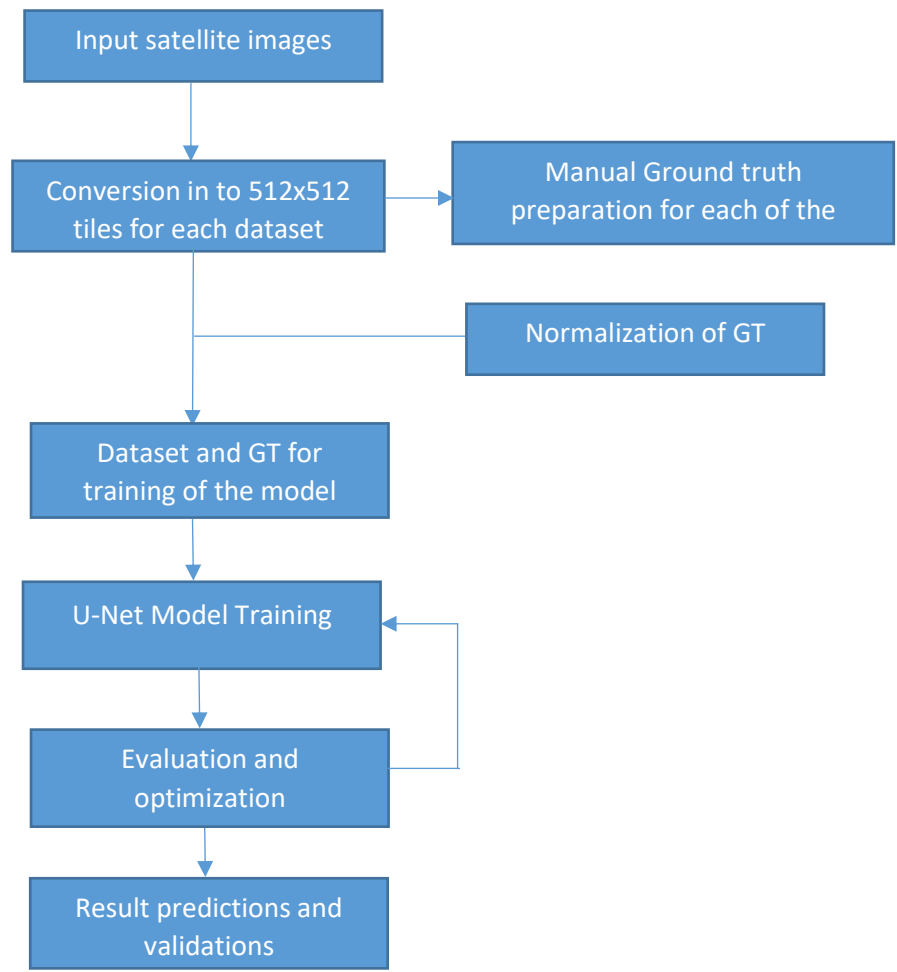

**Figure-2: Flowchart of Model processing.**

**VGG-UNet model architecture:**

VGG-UNet model was used in the problem of multi-label classification using Cartosat satellite data. Modified U-Net with five contractions and five expansion layers worked really well with binary segmentation problems whereas layers with VGG encoders were required to learn complex features present in the multi-label segmentation problem due to more complexity in the datasets. The left side of the encoder block had five blocks. It incorporated thirteen convolutional layers from the original VGG16 (Balakrishna, 2018). After each convolutional block, a max pooling operation which reduces the dimensions of the images by 2x2 was placed. The right side of the network is a decoder which also contained 5 blocks with 13 Convolutional layers. After each block, an up-sampling operation was kept to restore the dimensions of the image. The skip connections between the blocks were used to restore the dimensions of the image and was implemented using concatenate operation. Further, fully connected dense layer was for single dimension. In the last softmax activation layers was used to predict the segmented classes.

**Optimization, loss function and evaluation metrics:**

Optimization algorithms helps us to minimize an objective function $E(x)$ which is simply a mathematical function dependent on the model's internal learnable parameters which are used in computing the target values(Y) from the set of predictors(X) used in the model. There are popular methods like Gradient descent, Stochastic Gradient

Descent, Adam optimization methods in the deep learning. Adam is a combination of RMSprop and stochastic gradient decent with momentum algorithm. The method computes individual adaptive learning rates for different parameters from estimates of first and second moments of the gradients (Kingma D. and Jimmy L., 2017). Adaptive learning rate optimization (Adam) optimization function was used for our exercise as it gave advantage of dynamic handling of learning rate, took lesser time to converge and demonstrated better results.

A loss function—How the network will be able to measure its performance on the training data, and thus how it will be able to steer itself in the right direction. We used binary entropy function where classification was required in two categories whereas used categorical_ cross entropy function in multi-label classification. Cross entropy is defined as:

$$CE = -\sum_{i}^{C} t_i \log(s_i)$$ (Murphy K., 2012)

Where $t_i$ and $s_i$ are the ground truth and CNN score for each class I in C. in binary classification problem, where C' =2, cross entropy can be defined as

$$CE = -\sum_{i=1}^{C'=2} t_i \log(s_i) = -t_1 \log(s_1) - (1 - t_1)\log(1 - s_1)$$

Where it is assumed that there are two classes: $C_1$ and $C_2$. $T_1$ [0,1] and $s_1$ are the ground truth and score for $C_1$ and $t_2 = 1-t_1$ and $s_2 = 1-s_1$ are the ground truth and score for $C_2$.

**Evaluation metric:** For evaluation of the test results over unseen data, we used the Jaccard Index/ Intersection over Union (IoU) metric.

**Jaccard Index / IoU formula:**

The Jaccard index is a measure of the similarity between two sets of data. The higher the percentage, the more similar the two populations. The formula of Jaccard index is-

Jaccard Index = TP / (TP + FN + FP)   (Tan P. et al, 2006)

Here, TP, FP and FN are the total number of true positives, false positives and false negative pixels respectively.

## 4. Evaluation and results

As stated earlier, deep learning CNN models were trained and applied on Indian remote sensing datasets and objects such as buildings, ships and trees were automatically detected. Further, pixel level classification was performed using modified U-Net architecture as well as VGG-UNet architecture. Validation accuracy of better than 95% is obtained using IoU metric.

### 4.1 Building footprint detection:

U-net model was applied on CartoSat 1m imagery for extraction of building footprints. So, here pixel representing buildings were labelled as 0 and non-buildings are labelled as 1. 160 Images of 512x512 are supplied to train the model where an accuracy of >99% is achieved against the ground truth using IoU metrics. Further, model was applied on validation data (40 images) to understand the accuracy on unseen data. As, it is inferred from the below Figure-3(a),3(b), 3(c) and 3(d) that model predicted outputs are quite close to ground truths. Model correctly predicted the output results with an accuracy better than 95% as per the IoU accuracy metrics. Difference of ground truth and model output precisely depicts the accuracy of the model qualitatively.

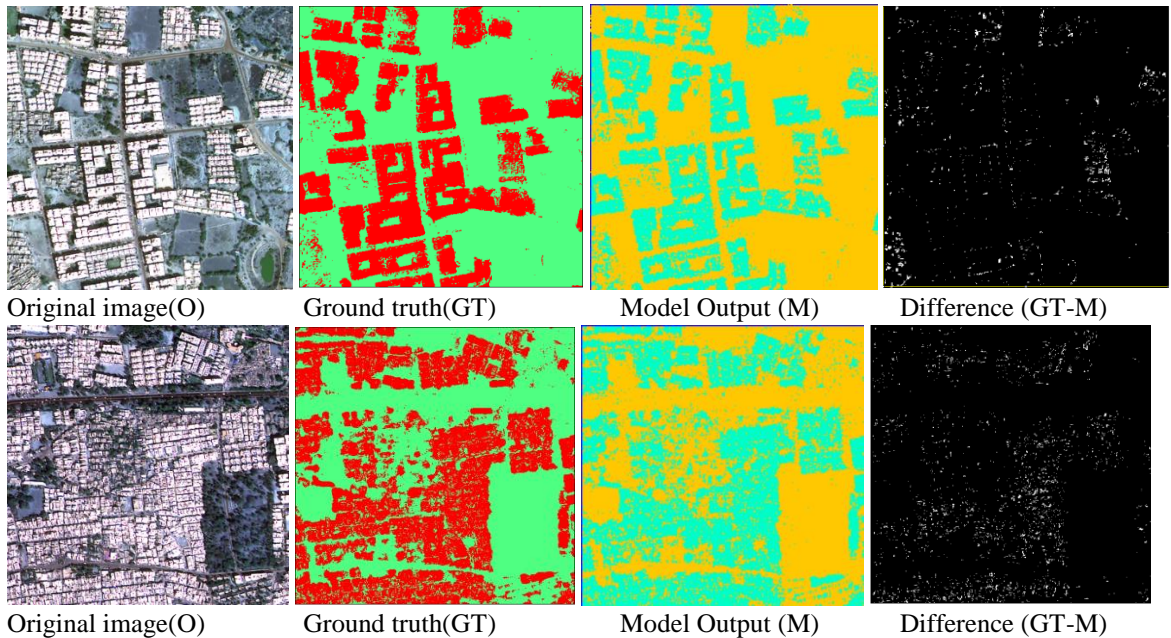

| Original image(O) | Ground truth(GT) | Model Output (M) | Difference (GT-M) |

**Figure: 3(a), 3(b), 3(c), 3(d), 3(e), 3(f), 3(g) and 3(h) represents Original image, ground truth, model predicted outputs and difference of GT and Model output respectively.**

### 4.2 Ship detection in optical data:

After successfully performing building detection, training and test dataset were prepared for the ship detection using CartoSat 1m. Same modified U-Net model architecture was trained on the separate dataset with an accuracy of 99.9% and validation accuracy (IoU) of more than 98% was achieved for ship detection. As, detection of ships was comparatively simpler task, better accuracy was obtained using the similar model. Results of Ship Detection on CartoSat dataset is shown in Figures 4(a), 4(b), 4(c) and 4(d) respectively. Difference of ground truth and model output precisely depicts the accuracy of the model qualitatively.

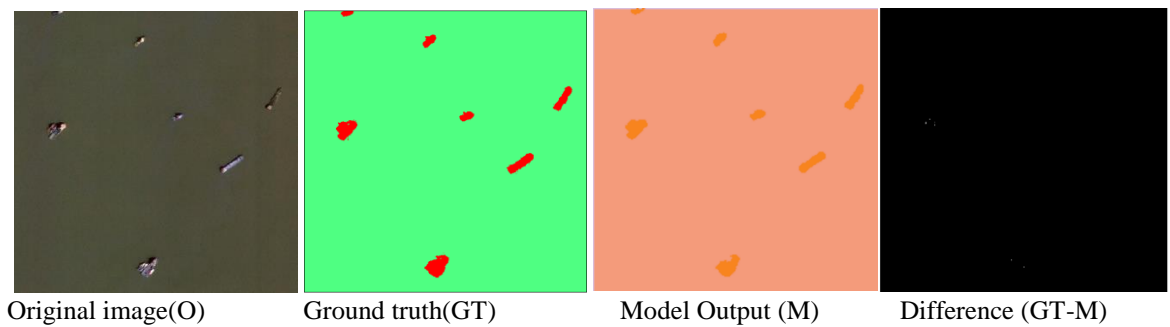

Original image(O)　　Ground truth(GT)　　Model Output (M)　　Difference (GT-M)

**Figure: 4(a), 4(b), 4(c) and 4(d) represents Original image, ground truth, model outputs for ship detection and difference of GT and Model output respectively.**

### 4.3 Tree detection in optical data:

For training and testing of this method, we utilized CartoSat2 series, CartoSat3 series satellite images and generated 8000 datasets, model is trained with 6000 training patches and 2000 patches for testing. Dimension of the patch size was 384x384. Training and test dataset were prepared for the tree detection using CartoSat. U-Net model architecture was trained on the dataset and Obtained IoU accuracy over built-up area is 91.6 percentages whereas IoU accuracy over field area is around 96.7 percent. Results of tree detection on CartoSat dataset is shown in Figure 5(a), 5(b), 5(c) and 5(d) respectively. Difference of ground truth and model output precisely depicts the accuracy of the model qualitatively.

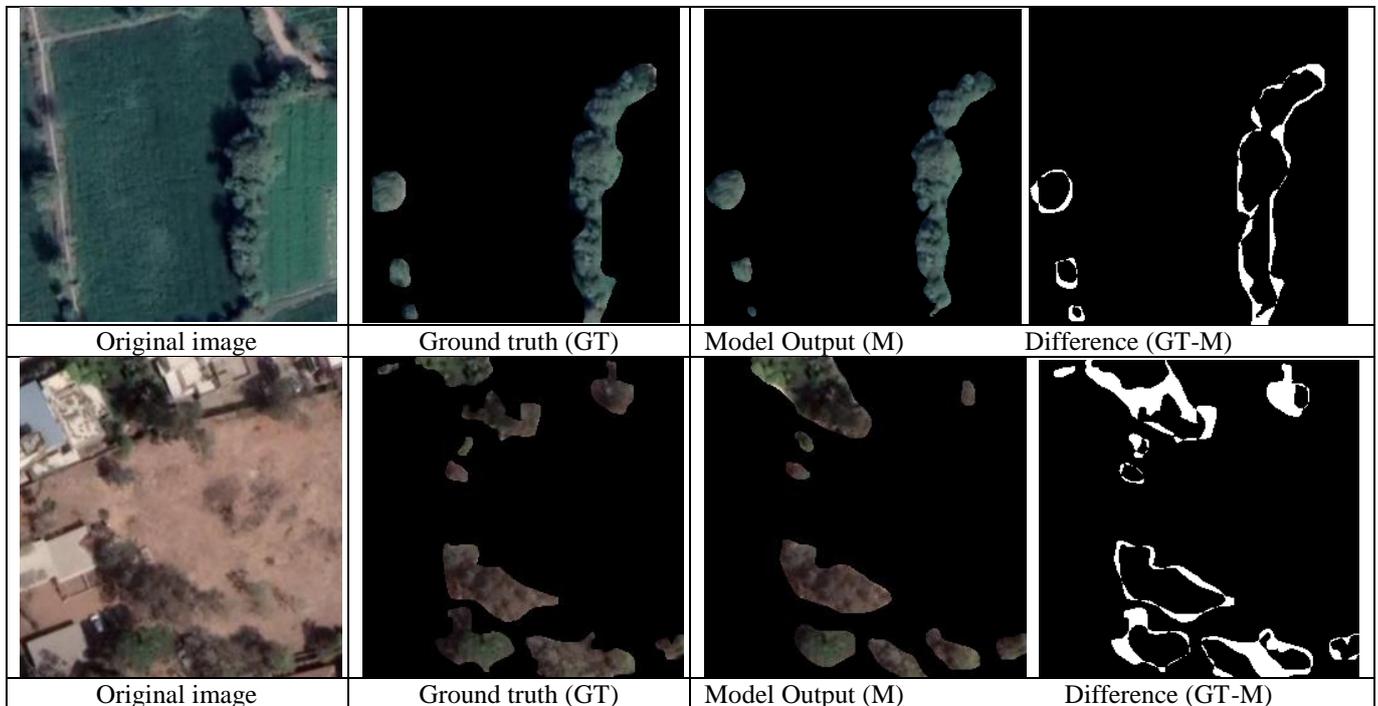

**Figure: 5(a), 5(b), 5(c), 5(d) represents original image, ground truth, outputs for Tree detection and difference of GT and Model output respectively.**

### 4.4 Ship detection using microwave datasets:

RISAT MRS data of 18m resolution was taken as an input dataset. Training data and test data was prepared separately for RISAT data using Data preparation and ground truth step. Training data and ground truth were fed into same U-Net model and ship detection was performed with an IoU accuracy of more than 99%. Results of Ship detection using microwave dataset is shown in Figures 6(a), 6(b), 6(c) and 6(d) respectively. Difference of ground truth and model output precisely depicts the accuracy of the model qualitatively.

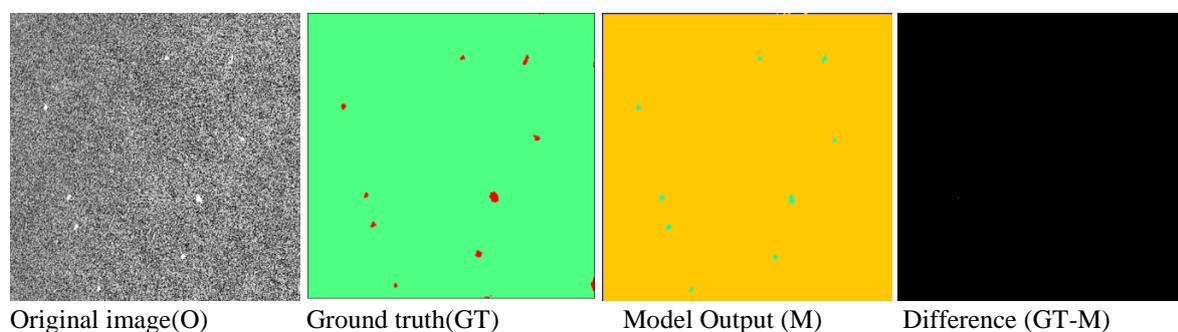

**Figure: 6(a), 6(b), 6(c) and 6(d) represents Original image, ground truth and predicted outputs for ship detection and difference of GT and Model output respectively in microwave dataset.**

4.5 **Tree detection using HR (1m) microwave datasets:**

High resolution microwave (1m) datasets was prepared along with the ground truth using 3000 images of 256x256 tile size. Training data and test data is selected as 80:20 ration and data was fed to UNet model for tree detection task. Obtained IoU accuracy over built-up area is 97 percentages whereas IoU accuracy over field area is around 85 percent. Results of tree detection on microwave dataset is shown in Figures 7(a), 7(b) 7(c) and 7(d) respectively. Difference of ground truth and model output precisely depicts the accuracy of the model qualitatively.

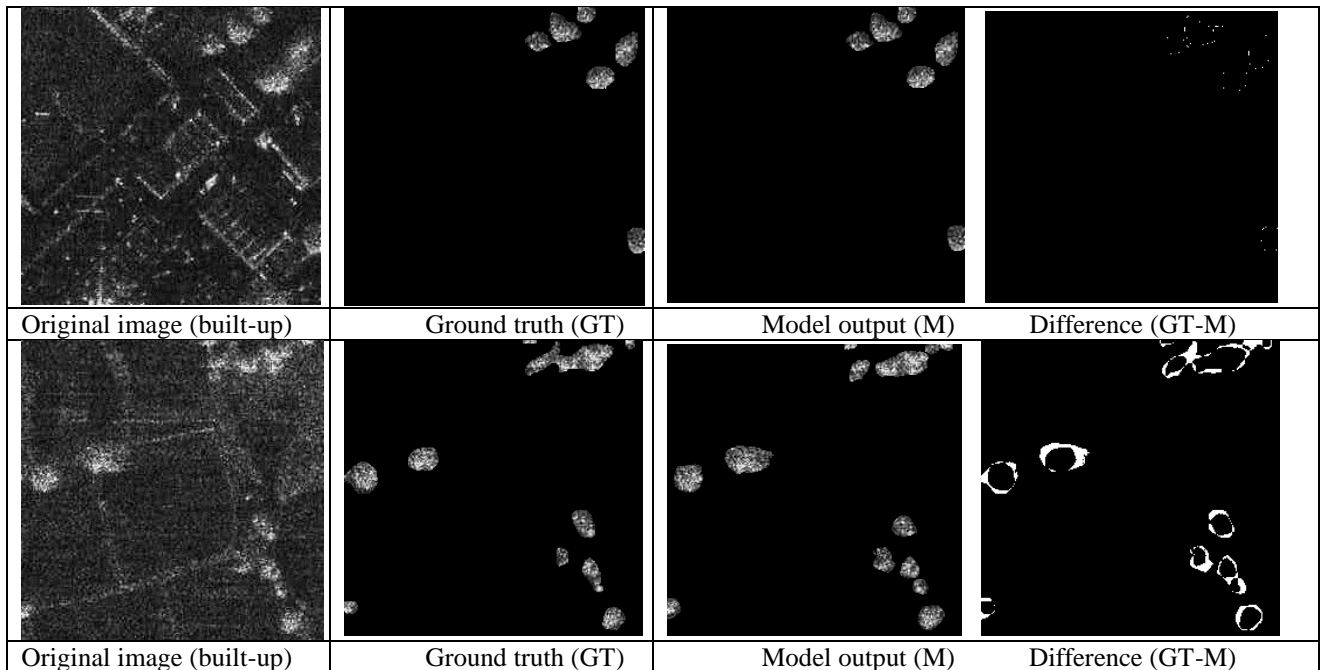

**Figure: 7(a), 7(b), 7(c), 7(d) represents Original image, ground truth, predicted outputs for tree detection in microwave dataset and difference of GT and Model output respectively in microwave dataset.**

### 4.6 Multi label classification results on Cartosat-2S (~1m) data

This was the most complex task among all. In this exercise, CartoSat multispectral dataset with a resolution of ~1m was used to classify classes like trees, urban structures, roads and barren land. Data was separately prepared manually by mapping input images to ground truth and sufficient number of 160 images were generated for Cartosat 1m images. Validation datasets of 40 images were created for testing the model accuracies. Our VGG-UNet model was trained on the new datasets and more than 99% training accuracy was achieved. Training and test accuracies of the model in all the cases were monitored after every epoch using callback mechanism. Here again, we got promising results with an accuracy (IoU) better than 95% as seen in Figure-8. Difference of ground truth and model output precisely depicts the accuracy of the model qualitatively.

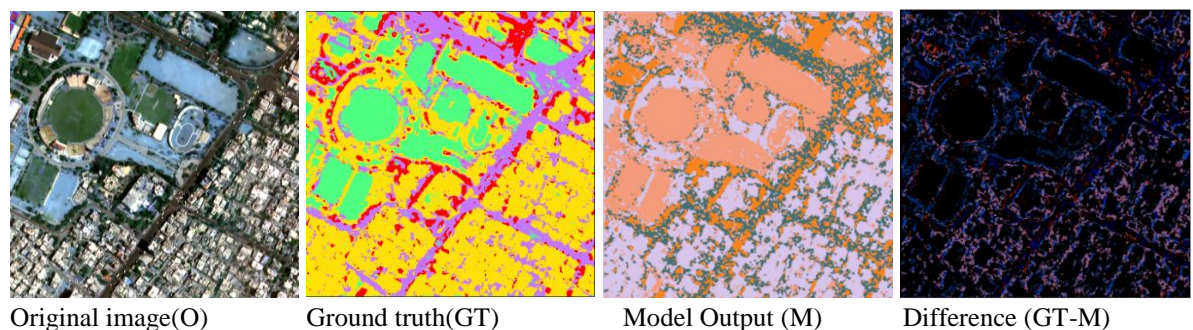

**Figure-8: Multi label classification results on Cartosat 1m data**

Overall, six different dataset banks were taken as an input, appropriate ground truth was prepared separately for each of the dataset, objects like buildings, ships, trees and multi-label segmentation were detected successfully using modified Unet and VGG-UNet based model architectures. Although, very good training accuracies were achieved using all the inputs but, IoU metric showed variation of results from 85% to 98% range (Table-2). There is fall in accuracies while predicting more complex features whereas single objects with less variations in the surrounding areas demonstrated highest accuracy. In problem-6 specially, homogenous regions exhibited almost 99% accuracy whereas differences are noted wherever existence of more complex features are seen or on the border areas of two features. With the preparation of more accurate ground truth data and availability of more datasets, better and more consistent results can be achieved.

| S. No. | Problem Definition | Deep learning model | Training accuracy | IoU validation accuracy |
|---|---|---|---|---|
| 1. | Building detection using high resolution optical data | Modified UNet | 99.9% | >95% |
| 2. | Ship detection using high resolution optical data | Modified UNet | 99.9% | >98% |
| 3. | Tree detection using high resolution optical data | Modified UNet | >95% | >91% |
| 4. | Ship detection using medium resolution microwave data | Modified UNet | 99.9% | >99% |
| 5. | Tree detection using high resolution microwave data | Modified UNet | >95% | >85% |
| 6. | Multi-label semantic segmentation using high resolution optical data | VGG-Unet | 99% | >95% |

**Table-2 Results of model accuracy on different problem types**

**Discussions**

Indian Space Research Organization has launched many remote sensing missions in the past for various earth observation studies. These missions have collected varied spatial resolution optical and microwave datasets. Due to high rate of data acquisitions, it is essential to automatically process and to detect objects such as buildings/ ships/ trees for various applications. Using deep learning models and Semantic segmentation technique, accurate detection and classification of images are possible. Many researcher across the world worked on semantic segmentation using open datasets or remote sensing data from contemporary missions. Prathap et al. (2018) used open data from SpaceNet and performed building detection with an accuracy of 0.58 to 0.88 on various datasets using U-Net model. Ivanovsky et al. (2019) used 14 color 0.5m aerial images for detection of building with an accuracy of 96.31 %. Zhou et. al. (2019) worked on ship detection from PolSAR images based on CNN. Based on figure of merit a metric, their model demonstrated accuracy of 85-100% depending different datasets. C. Gonzales et al. (2019) did semantic segmentation of clouds in satellite imagery using pre-trained U-Net models with an accuracy of 86.19% on Jaccard Index. Ulmas et al. (2020) used modified U-Net model for creating land cover classification mapping with a F1 score of 0.749 on multiclass land cover by utilizing sentinel-2 datasets. In some latest developments, Guerin et al (2021) proposed a method for automatic semantic segmentation of satellite image in to six classes with mIoU of 54.22 using open data provided by French department. Many researchers attempted U-Net model and semantic segmentation technique in the past but, our modified UNet model is unique in such a way that it converges very quickly on fewer datasets and demonstrated high accuracies. Although, model was trained separately on each data bank but, same model architecture fits well to detect ships/ buildings / trees from varied resolution microwave and optical imagery. VGG-Unet based model also performed really well on for multi-label segmentation and achieved IoU accuracy of better than 95%. In this research work, we solved six different problem using modified UNet model and VGG UNet model and proved that promising results can be achieved using the approach of deep learning over varied resolution satellite datasets.

**Conclusions**

Deep learning models were widely used to solve many problems in image processing. In this research, we worked on semantic segmentation and its applications on Indian remote sensing datasets using deep learning. Optical and microwave datasets were used for detection of buildings, ships, trees and multi-label classifications in the various exercises presented in this paper. Deep learning models based on modified U-Net architecture and VGG-Unet architecture were trained for each datasets against the ground truth and results were generated. Results were validated with IoU metrics for all the exercises and a consistent more that 95% accuracy is obtained over unseen data as shown in Table-2. By conducting all the exercises in this paper, we conclude that semantic segmentation shows good promising results on satellite imageries of multi-sensor and varied spatial resolutions (2m to 18m). Technique is very successful to automatically detect / classify various objects in the satellite data. Semantic segmentation has many future applications relevant for diverse domains such as automatic detection of different features from satellite images/ aerial images, automatic filtering of feature based data, automatic change detection and instance segmentation.

**Funding Statement:** No funding was received for this study.

**Competing Interest Statement**: The authors declare they have no competing interests.

**Data Availability Statement (DAS):** Data used for this study is restricted in nature.

**Code Availability:** Custom code is also restricted in nature.

**Author Contributions statement:** "All authors contributed to the study conception and design. Material preparation, data collection and analysis were performed by Jai G Singla and Bakul Vaghela. Semantic segmentation work on building detection, ship detection and multi-label classification is carried out by Jai G Singla whereas work on tree detection in optical and microwave datasets is carried out by Bakul Vaghela . The first draft of the manuscript was written by Jai G Singla. All authors read and approved the final manuscript."

**Acknowledgments**

The authors express sincere gratitude to Shri. N. M. Desai, Director, Space Applications Centre for his permitting the presentation of this paper. Suggestions from internal referees to improve an earlier version of this paper are sincerely acknowledged.